\documentclass[11pt,a4paper]{article}
\usepackage[hyperref]{acl2020}
\usepackage{times}
\usepackage{latexsym}
\usepackage{hyperref}
\usepackage{graphicx}
\usepackage{multirow}

\usepackage{microtype}

\aclfinalcopy % Uncomment this line for the final submission
\title{Measuring Alignment Bias in Neural Seq2Seq Semantic Parsers}

\author{Davide Locatelli \\
  Technical University of Catalonia \\
  Barcelona, Spain \\
  \texttt{davide.locatelli@upc.edu} \\\And
  Ariadna Quattoni \\
  Technical University of Catalonia \\
  Barcelona, Spain \\
  \texttt{aquattoni@cs.upc.edu} \\}

\date{}

\begin{document}
\maketitle
\begin{abstract}
Prior to deep learning the semantic parsing community has been interested in understanding and modeling the range of possible word alignments between natural language sentences and their corresponding meaning representations. Sequence-to-sequence models changed the research landscape suggesting that we no longer need to worry about alignments since they can be learned automatically by means of an attention mechanism. More recently, researchers have started to question such premise. In this work we investigate whether seq2seq models can handle both simple and complex alignments. To answer this question we augment the popular \textsc{Geo} semantic parsing dataset with alignment annotations and create \textsc{Geo-Aligned}. We then study the performance of standard seq2seq models on the examples that can be aligned monotonically versus examples that require more complex alignments. Our empirical study shows that performance is significantly better over monotonic alignments. \footnote{The code and data is publicly available at \url{https://github.com/interact-erc/geo-aligned}}
\end{abstract}

\section{Introduction}

In semantic parsing, the goal is to map natural language (NL) sentences into machine-readable meaning representations (MR) which allow for automated reasoning. For example, consider the following pair:\\

NL : {\it What is the population of Georgia ?}

MR : {\it answer (population (state (georgia) ) )}\\

Prior to deep learning models, a popular approach was to learn a grammar-based parser that explicitly models alignments between the NL and MR sequences \cite{wong2006, zettlemoyer2005, zettlemoyer2007, lu2008, kwiatkowksi2010, kwiatkowski2011}. The emergence of sequence-to-sequence (seq2seq) semantic parsers with attention mechanisms changed the research landscape: one of the initial premises of seq2seq models is that alignments no longer need to be explicitly modeled because the attention mechanisms will automatically learn them \cite{bahdanau2015}. More recently, researchers started to question such premise, having observed that seq2seq models fail to make proper generalizations on out-of-distribution test sets on which traditional grammar-based models excel \cite{liu2020, liu2021, wang2021}. 

In this paper we follow this line of research and ask the questions: Can standard seq2seq models handle arbitrary alignments? And if not, what kind of alignment bias do they have? To answer these questions, we augment the \textsc{Geo} semantic parsing benchmark \cite{zelle1996} with alignment annotations and create \textsc{Geo-Aligned}. We then compare the performance of seq2seq models on examples that can be easily aligned with simple monotonic alignments to the performance of these models on examples that require word reordering. Our empirical study shows that seq2seq parsers perform significantly better over examples that can be monotonically aligned. In other words, the flexibility of not having to explicitly model alignments comes at a cost: seq2seq models have difficulties in learning complex alignments.

The main contributions of this paper are: 
\begin{enumerate}
    \item We introduce a new dataset: \textsc{Geo-Aligned} that augments the \textsc{Geo} semantic benchmark with alignment annotations. We used the English and German versions of the original dataset, and we additionally introduce a new Italian version.
    \item Using \textsc{Geo-Aligned} we define new evaluation splits to distinguish parsing performance over easier and harder examples.
    \item Our empirical study shows that seq2seq parsers are significantly better in handling monotonic alignments, and quantifies the impact of using attention.
    \item As a side contribution we offer a measure of the complexity of the \textsc{Geo} dataset, showing that more than half of the examples involve monotonic alignments.
\end{enumerate}

\section{The \textsc{Geo-Aligned} Benchmark}
In this section we describe the \textsc{Geo-Aligned} dataset, an augmentation of the popular \textsc{Geo} semantic parsing benchmark first introduced by \citet{zelle1996}. We start by providing a brief formal definition of word alignments following standard notation from the statistical machine translation literature, and we define monotonic and non-monotonic alignments \cite{dekaiwu}. We then detail how we augment the \textsc{Geo} dataset and provide statistics that measure the complexity of the dataset.

\subsection{Bi-text alignments}
Given an input sequence of $N$ words ${\bf x} = x_1, \ldots, x_{N}$, and a target sequence of $M$ words ${\bf y} = y_1, \ldots, y_{M}$, a bi-text is defined as the tuple $({\bf x}, {\bf y})$. A bi-text word alignment is a set of bi-symbols $\mathcal{A}$, where each bi-symbol $(x_i,y_j)$ couples a word $x_i$ in the input sequence at position $i$ to a word $y_j$ in the target sequence at position $j$. 

If a word $x_i$ from the input sequence does not need an alignment to a word in the target, we introduce an $\varepsilon$ in ${\bf y}$ at position $i$. This bi-symbol $(x_i, \varepsilon_i)$ amounts to a deletion, i.e. mapping from input to target involves deleting a word from the input. Conversely, if a word $y_j$ from the target does not require an alignment to a word in the input, we introduce an $\varepsilon$ in ${\bf x}$ at position $j$. This bi-symbol $(\varepsilon_j, y_j)$ amounts to an insertion, i.e. mapping from input to target involves inserting an extra word in the target. We refer to the number of insertions and deletions in an alignment as the gap length. Figure \ref{fig:example-alignments} shows examples of alignments from the \textsc{Geo-Aligned} dataset.

\subsection{Monotonic and non-monotonic alignments}
\begin{figure}
\centering
\includegraphics[width = 1\hsize]{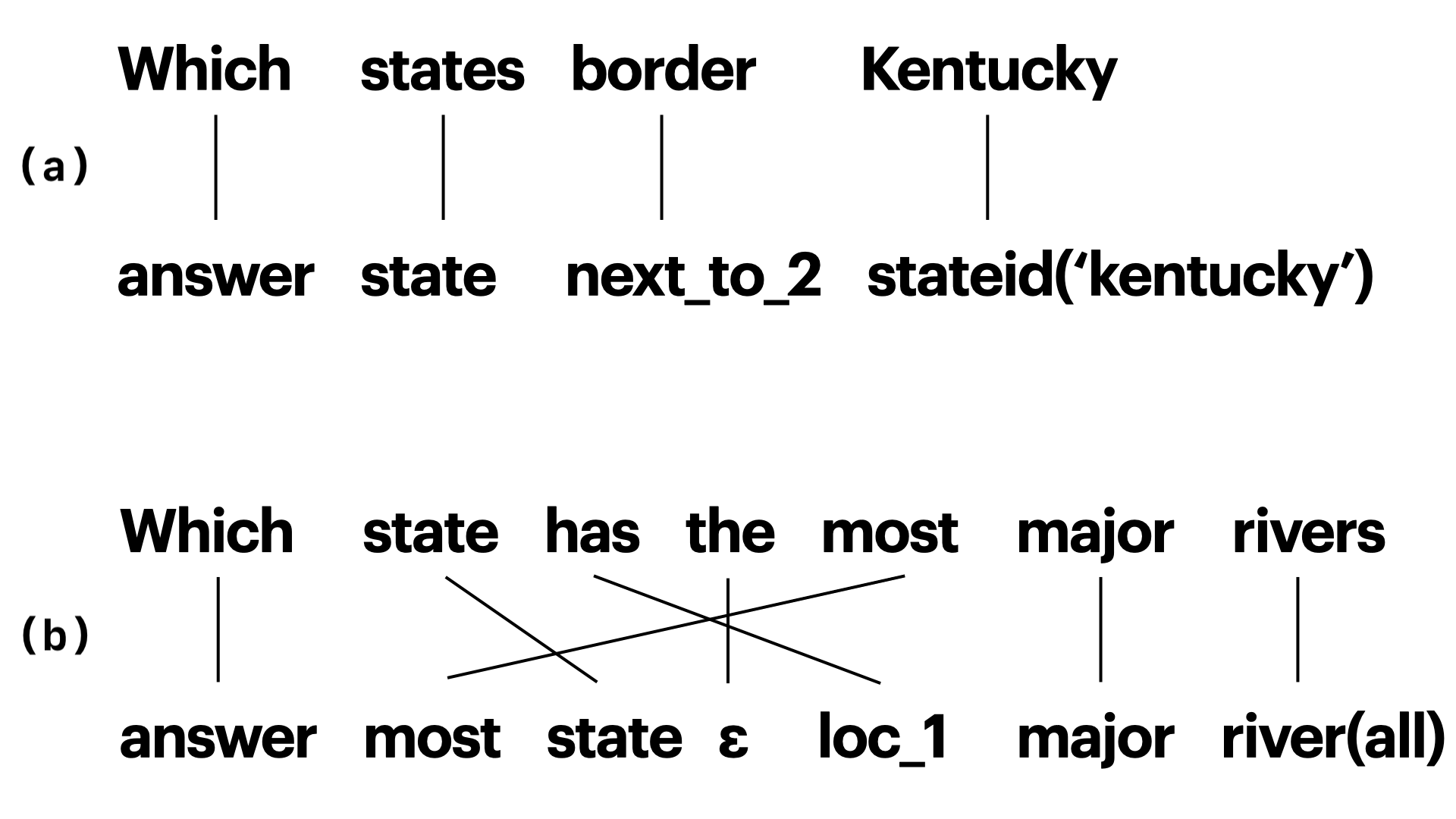}
\caption{Examples alignments from the \textsc{Geo-Aligned} benchmark. Each bi-symbol is represented as a vertical line coupling words in the NL with words in the corresponding MR. The monotonic alignment (a) does not involve crossings of bi-symbols, while the non-monotonic alignment (b) involves considerable reordering.}
\label{fig:example-alignments}
\end{figure}

Monotonic alignments are bi-text alignments where $\mathcal{A}$ contains bi-symbols of the forms $(x_i, y_j)$, $(x_i, \varepsilon_j)$ or $(\varepsilon_i, y_j)$ where $i=j$. In other words, a monotonic alignment does not involve any reordering of the words. Conversely, non-monotonic alignments also include bi-symbols of the form $(x_i, y_j)$ where $i \neq j$. Figure \ref{fig:example-alignments} shows an example of a monotonic alignment versus a non-monotonic one.

\subsection{Alignment annotation}

The original \textsc{Geo} dataset contains 880 English questions about US geography, paired with a meaning representation. Several MR formalisms have been introduced for this dataset, including a first-order logic as in \citet{zelle1996}, a variable-free functional language introduced by \citet{kate2005} and SQL \cite{popescu2003, giordani2012, iyer2017}. In \textsc{Geo-Aligned}, we use the variable-free functional language formalism. Similarly to \citet{wang2021}, we further simplify the MR by removing the brackets. This is done to avoid introducing numerous $\varepsilon$ in the alignments, and also to better reveal the structural similarity between the NL and MR sequences. Similarly to \citet{dong2016}, we remove constants used to identify states, rivers, cities, places and countries by substituting them with their type.

Alignments were provided by four expert annotators. For each pair, the annotators were first asked to decide whether there was a monotonic or non-monotonic alignment. Secondly, annotators were asked to provide the actual alignment from NL to MR words. More specifically, two annotators aligned the entire dataset, while the other two each annotated fifty disjoint examples. Inter-annotation agreement was calculated by comparing the alignments provided. A first agreement metric is Cohen's Kappa statistic \cite{cohen1960} to measure the agreement of monotonic versus non-monotonic labels: the average score obtained is 0.803, which corresponds to substantial agreement. We then calculated the average percentage of exact matches between the alignments of the two main annotators and each of the other three, which resulted in a 90\% average match. Disagreements were resolved by keeping the annotation that best matched the alignment strategy taken by the majority.

Bi-text word alignments vary depending on the order in which the words appear both in the natural language and the meaning representation \cite{steedman2020}. If we keep the MR fixed, a sentence in one language might be monotonically aligned, while the same sentence in another language might not be. To better understand the range of alignments between natural language utterances and meaning representations one should ideally consider multiple languages. With this objective in mind, we additionally annotated the German version \cite{jones2012} of \textsc{Geo}, and a new Italian version that we introduce, obtained by translations of the English sentences provided by an Italian native speaker.

The resulting dataset contains the NL and MR data pairs, augmented with 
\begin{itemize}
    \item a label indicating whether there is a monotonic alignment;
    \item the alignment that maps NL and MR words.
\end{itemize}
Table \ref{tab:alignment_stats} reports annotation statistics for \textsc{Geo-Aligned}. In general, it can be observed that across all languages the majority of the alignments are monotonic and the average gap length is less than three. For non-monotonic alignments the average number of reordered words is below three. 

\begin{figure}
    \centering
    \includegraphics[width=0.99\linewidth]{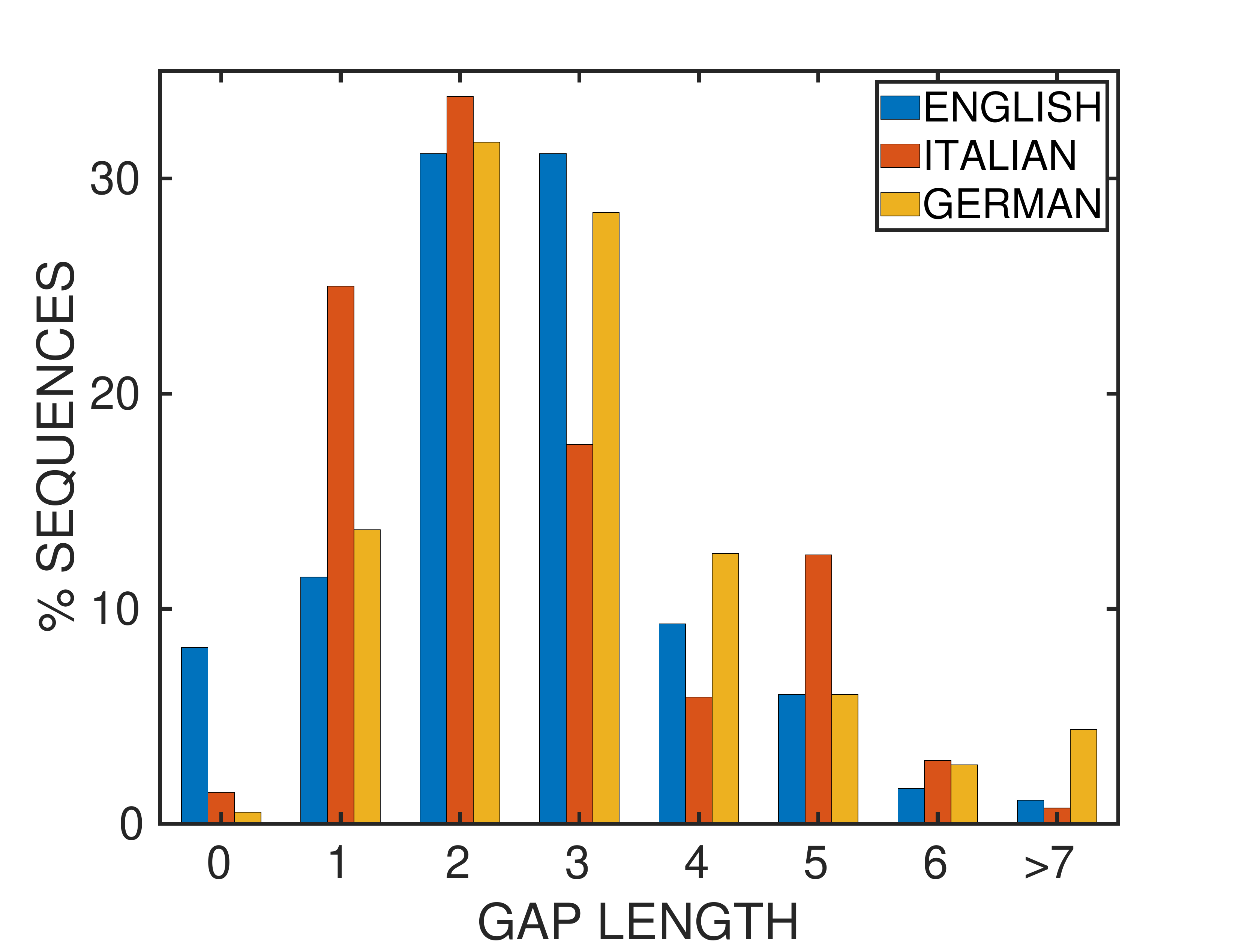} 
    \caption{Distribution of gap lengths for the monotonic alignments.}
    \label{fig:gap_histogram}
  \end{figure}

\begin{table}[ht]
    \centering
    \begin{tabular}{llllll}
    \hline
    {\bf Lang} & {\bf Len} & {\bf MP} & {\bf MG} & {\bf M0} & {\bf NMR} \\
    \hline 
    EN & 7.67 & 0.75 & 2.52 & 8.2 & 2.14 \\ 
    DE & 7.72 & 0.65 & 2.91 & 0.55 & 2.52 \\ 
    IT & 7.92 & 0.52 & 2.54 & 1.5 & 2.23 \\ 
\hline 
\end{tabular}
\caption{Alignment annotation statistics for different languages. Len is the mean length of input NL sentences, MP is the percentage of monotonic alignments, MG is the average gap in monotonic alignments, M0 is the percentage of monotonic alignments with no gap, and NMR is the average number of words reordered in the non-monotonic alignments.}
\label{tab:alignment_stats}
\end{table}
With respect to differences between the three languages, Figure \ref{fig:gap_histogram} shows a histogram of the gap lengths of monotonic alignments. As we can see the distributions are quite similar, but slightly shifted towards longer gaps for German and Italian. In particular, there are significantly more alignments with no gap in English. The proportion of monotonic alignments reflects the structural similarity between the variable-free MRs and the NL sequences. It is highest in the case of English, after which the MR formalism was modeled. German is syntactically more similar to English than Italian and as a result it can be more easily aligned with the MR sequences. An exemplary syntactic difference is adjective placement: in English and German adjectives come before nouns, whilst in Italian they are usually placed after. When a superlative is used in the NL sentence, the MR, being modeled after English, places it before the noun. This creates a monotonic alignment with English and German sentences and a non-monotonic one with Italian ones. For example, if the question is {\it What is the largest state ?} the corresponding MR will be {\it answer(largest(state(all)))}. Because {\it largest} comes before {\it state} in both English and German as well as in the MR, the alignment will be monotonic. In Italian, {\it largest} comes after {\it state} and the alignment will require reordering.

\section{Measuring Alignment Bias}
\subsection{Models and Experiments}
The goal of our study is to compare the performance of neural seq2seq models over monotonic and non-monotonic alignments. Our hypothesis is that seq2seq models can implicitly learn monotonic alignments more easily than non-monotonic alignments. To evaluate this hypothesis we compared the performance of two seq2seq architectures on \textsc{Geo-Aligned}.\\

\noindent {\bf \textsc{LSTM Seq2Seq}} A standard seq2seq model based on a bidirectional-LSTM encoder \cite{hochreiter1997, schuster1997}, and a unidirectional LSTM decoder that uses attention \cite{bahdanau2015}. We then ablate the decoder of the attention layer to investigate its impact on the performance for the different alignments.\\

\noindent {\bf \textsc{BART}} A pre-trained seq2seq model based on a bidirectional encoder and a left-to-right decoder \cite{bart}. Since it was pre-trained on English corpora, we only used this model on the English version of the dataset.\\

For our experiment we use exact-match accuracy as the evaluation metric, i.e. the percentage of exact matches between the predicted and ground-truth MRs. The alignment labels in \textsc{Geo-Aligned} allow us to break down the accuracy score for the two classes of alignments and observe whether the seq2seq framework has an implicit bias towards monotonic alignments. Further implementation and experimental setup details can be found in Appendix \ref{sec:appendix}.

\subsection{Results}
Table \ref{tab:results} shows the performance for the different models and languages. As we can observe accuracy for all models is significantly lower over non-monotonic alignments and this is true for all languages. The difference in performance between monotonic and non-monotonic alignments is more pronounced for models with no attention, but it holds true for all of them.

\begin{table}
    \centering
    \begin{tabular}{lllll}
    \hline
    {\bf Lang} & {\bf Model} & {\bf Acc} & {\bf MAcc} & {\bf NMAcc} \\ 
    \hline 
    \multirow{3}{2em}{\textsc{En}} & $\textsc{Lstm}$ & 0.83 & 0.87 & 0.74 \\ 
    & $\textsc{Lstm}$-attn & 0.75 & 0.80 & 0.61 \\
    & $\textsc{BART}$ & 0.85 & 0.87 & 0.80 \\
    \hline
    \multirow{2}{2em}{\textsc{De}} & $\textsc{Lstm}$ & 0.63 & 0.73 & 0.54 \\ 
    & $\textsc{Lstm}$-attn & 0.57 & 0.69 & 0.46 \\
    \hline
    \multirow{2}{2em}{\textsc{It}} & $\textsc{Lstm}$ & 0.77 & 0.84 & 0.71 \\ 
    & $\textsc{Lstm}$-attn & 0.71 & 0.80 & 0.63 \\
    \hline
\end{tabular}
\caption{Summary of results for the different models and languages: LSTM is the seq2seq model based on a bidirectional LSTM encoder and an LSTM decoder with attention. LSTM-attn ablates the attention layer in the decoder. Acc reports the overall accuracy for each model, MAcc and NMAcc are the accuracy over sequences with monotonic and non-monotonic alignments respectively.}
    \label{tab:results}
\end{table}

\begin{figure}
    \centering
    \includegraphics[width=0.99\linewidth]{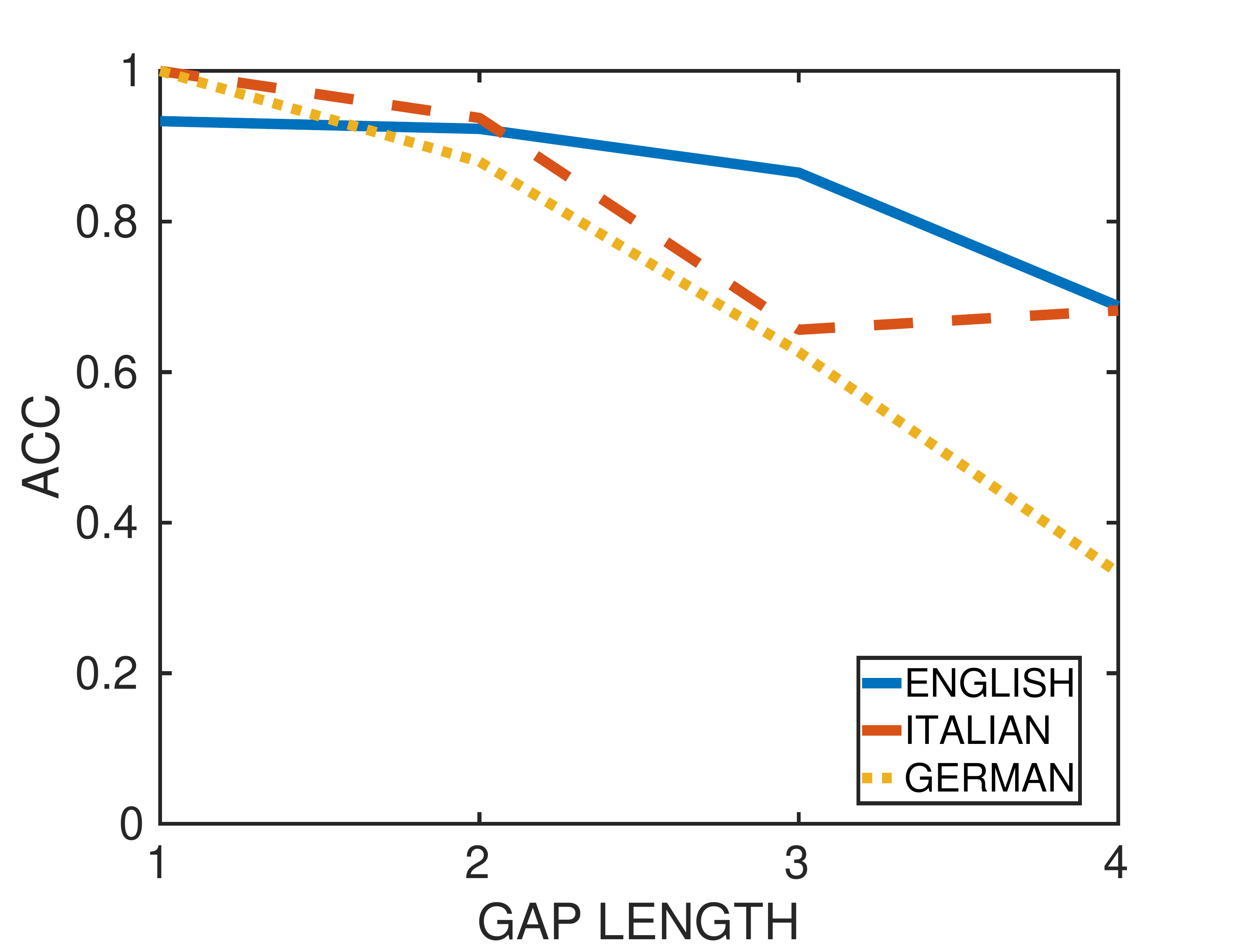} 
    \caption{Accuracy for monotonic examples as a function of gap length.}
    \label{fig:acc_vs_gap}
\end{figure}

The performance follows the same pattern across languages and models: accuracies are higher for monotonic sequences than for non-monotonic ones. For English and Italian the differences are quite similar: models with attention score 0.13 point higher for monotonic sequences; without attention the difference is 0.19 for English and 0.17 for Italian. German has a lower accuracy overall. One possible explanation (as shown in Figure \ref{fig:gap_histogram}) is that the monotonic gap distribution for these two languages has a slight shift towards shorter gaps and in particular the sequences with no gap could help the models to implicitly induce better alignments. Moreover, the difference between monotonic and non-monotonic performance is starker: the model scored 0.19 and 0.23 better on monotonic examples with and without attention respectively. This might be due to the fact that more words are reordered on average for German than for the other two languages (see Table \ref{tab:alignment_stats}). Figure \ref{fig:acc_vs_gap} shows accuracy for monotonic sequences binned by gap length. We observe that for all languages there is a negative correlation between accuracy and gap length.

We performed a qualitative analysis of the predictions by categorizing errors based on how many steps are needed to correct the mistake. Simpler errors are those where the correct MR can be recovered by inserting, deleting or changing at most two tokens, without reordering. An example is:\\

MR: {\it answer river loc$\_$2 stateid state$\_$name}

prediction: {\it answer loc$\_$2 stateid state$\_$name}\\

where the gold MR can be recovered by inserting {\it river} in the second position. More complex errors require correcting three or more tokens, and can also require reordering of the output. Table \ref{tab:qualitative} reports statistics of our analysis. In general, we found that errors on monotonic examples are of the simpler category in much higher proportion than for non-monotonic: across languages, non-monotonic sequences require much more complex corrections involving three or more tokens as well as considerable reordering. 

Another interesting finding is that, despite BART and our LSTM-based seq2seq model achieve similar results in English (see  Table \ref{tab:results}), the LSTM-based model makes more complex mistakes, particularly in the monotonic case. For these examples, the vast majority of the errors for BART were one-token, and we found that most of these were minor mistakes such as predicting the token $loc\_2$ instead of $loc\_1$. The predictions of the LSTM-based model are more dissimilar to the gold MR.

\begin{table}
    \centering
    \begin{tabular}{llllll}
    \hline
    {\bf Lang} & {\bf Align} & {\bf Model} & {\bf 1T} & {\bf 2T} & {\bf Other} \\ 
    \hline 
    \multirow{4}{2em}{\textsc{En}} & M & $\textsc{Lstm}$ & 0.46 & 0.19 & 0.32 \\
    &NM & $\textsc{Lstm}$ & 0.24 & 0.15 & 0.61 \\
    &M & $\textsc{BART}$ & 0.67 & 0.25 & 0.08 \\
    &NM & $\textsc{BART}$ & 0.29 & 0.17 & 0.54 \\
    \hline
    \multirow{2}{2em}{\textsc{De}} & M & $\textsc{Lstm}$ & 0.72 & 0.08 & 0.20 \\
    &NM & $\textsc{Lstm}$ & 0.32 & 0.27 & 0.41 \\
    \hline
    \multirow{2}{2em}{\textsc{It}} & M & $\textsc{Lstm}$ & 0.72 & 0.05 & 0.23 \\
    &NM & $\textsc{Lstm}$ & 0.43 & 0.18 & 0.39 \\
    \hline
\end{tabular}
\caption{Statistics of qualitative analysis on prediction errors. Align indicates the type of alignment: M stands for monotonic, NM for non-monotonic. 1T is the proportion of examples requiring a one-token correction without reordering. Similarly, 2T is for two-token corrections without reordering. Other is the proportion of examples requiring more complex corrections of three or more tokens, occasionally with reordering.}
    \label{tab:qualitative}
\end{table}

\section{Related Work}

Several grammar formalisms have been proposed for semantic parsing, including categorical grammars \cite{steedman1996,steedman2000, zettlemoyer2005,clark2003,zettlemoyer2007, kwiatkowksi2010, kwiatkowski2011} and synchronous context free grammars \cite{wong2006}. Both approaches model alignments explicitly and they are induced from data. There have also been attempts to derive a more general formalism to unify the different grammar based approaches to semantic parsing \cite{jones2011}.

More recently, neural seq2seq models were proposed for semantic parsing in \citet{dong2016, jia2016, iyer2017}. The seq2seq approach aims to relax the reliance upon high-quality lexicons, i.e. domain-specific word alignments. Most seq2seq systems implement an attention mechanism such as those proposed by \citet{bahdanau2015, luong2015, xu2015}, which can be seen as a strategy to learn soft alignments \cite{dong2016}.

Recently there has been an interest in testing the generalization abilities of neural semantic parsers, which resulted in the creation of several new benchmarks \cite{bastings2018, lake2018, loula2018, ruis2020, keysers2020, kim2020} on which recent work has shown improved performance by introducing more alignment bias in the models either explicitly \cite{liu2021}, or implicitly \cite{wang2021}.

\section{Conclusion}
In this paper we introduced the \textsc{Geo-Aligned} dataset that offers an evaluation framework for testing the performance of semantic parsers over examples of varying alignment complexity. Our experiments have shown that seq2seq neural parsers perform significantly better over simpler monotonic alignments, suggesting that they have an implicit bias. We hope that \textsc{Geo-Aligned} can be used by other researchers to further test alignment biases. 

\section*{Acknowledgments}
We thank the anonymous reviewers for their valuable feedback, as well as the other members of the \textsc{INTERACT} group at Universitat Politècnica de Catalunya. This work is supported by the European Research Council (ERC) under the European Union’s Horizon 2020 research and innovation program (grant agreement No.853459 ). The authors gratefully acknowledge the computer resources at Artemisa, funded by the European Union ERDF and Comunitat Valenciana, as well as the technical support provided by the Instituto de Fisica Corpuscular, IFIC (CSIC-UV).

\bibliography{acl2020}
\bibliographystyle{acl_natbib}

\appendix

\section{Implementation and training details}
\label{sec:appendix}
We based our LSTM-based seq2seq model on \citet{bahdanau2015}. We use a one-layer bidirectional LSTM for our encoder and a one-layer unidirectional LSTM for our decoder. At training we minimize the cross entropy loss between the predictions and the ground-truth MR sequences. We use a batch size of 32, Adam optimizer and learning rate of 0.001. We manually tune the hyperparameters, and train for 100 epochs on one NVIDIA TESLA V100 16GB GPU.

For BART, we used the pre-trained BART-base model provided by the HuggingFace transformers library \cite{huggingface}. We fine-tune for 100 epochs with a learning rate of 0.00001 on one NVIDIA TESLA V100 16GB GPU. Fine-tuning took approximately 1h30mins.
\end{document}